\documentclass[journal]{IEEEtran}

\ifCLASSINFOpdf
\else
   \usepackage[dvips]{graphicx}
\fi
\usepackage{url}

\usepackage{graphicx}

\usepackage{amsmath,multirow}
\usepackage{graphicx}
\usepackage{algorithm}
\usepackage{algorithmic}
\usepackage{subfigure,color}

\usepackage{gensymb,graphics,subfigure,inputenc}
\graphicspath{{Figures/}}
\usepackage{textcomp,subfigure,multirow,upgreek}
\usepackage{booktabs}
\usepackage{color,mdwlist}
\usepackage{lineno,hyperref}
\usepackage{color}
\usepackage{soul}

\usepackage{amsfonts}
\usepackage{amsmath}
\usepackage{amssymb}
\usepackage[table]{xcolor}

\usepackage{tikz}
\usepackage{bbding}
\usepackage{pifont}
\usepackage{wasysym}
\usepackage{amssymb}

\begin{document}

\title{Modiff: Action-Conditioned 3D Motion Generation with Denoising  Diffusion Probabilistic Models}

\author{Mengyi Zhao, Mengyuan Liu, Bin Ren, Shuling Dai, and Nicu Sebe 
\thanks{M. Zhao is with the State Key Laboratory of Virtual Reality Technology and Systems, Beihang University, Beijing 100191, China (e-mail: zhaomengyi@buaa.edu.cn).}
\thanks{Mengyuan Liu is with the School of Intelligent Systems Engineering, Sun Yatsen University, Shenzhen 510275, China, and also with Guangdong Provincial Key Laboratory of Fire Science and Technology, Guanzhou 510006, China (e-mail: liumy85@mail.sysu.edu.cn).}
\thanks{B. Ren is with University of Pisa \& University of Trento, Italy (e-mail: bin.ren@unitn.it).}
\thanks{S. Dai is with the State Key Laboratory of Virtual Reality Technology and Systems, Beihang University, Beijing 100191, China, and also with Jiangxi Research Institute, Beihang University, China (e-mail: sldai@buaa.edu.cn).}
\thanks{N. Sebe is with University of Trento, Trento, Italy (e-mail:niculae.sebe@unitn.it).}}
 
\markboth{}
{Shell \MakeLowercase{\textit{et al.}}: Bare Demo of IEEEtran.cls for IEEE Journals}
\maketitle
\begin{abstract}
Diffusion-based generative models have recently emerged as powerful solutions for high-quality synthesis in multiple domains. Leveraging the bidirectional Markov chains, diffusion probabilistic models generate samples by inferring the reversed Markov chain based on the learned distribution mapping at the forward diffusion process. In this work, we propose Modiff, a conditional paradigm that benefits from the denoising diffusion probabilistic model (DDPM) to tackle the problem of realistic and diverse action-conditioned 3D skeleton-based motion generation. We are a pioneering attempt that uses DDPM to synthesize a variable number of motion sequences conditioned on a categorical action. We evaluate our approach on the large-scale NTU RGB+D dataset and show improvements over state-of-the-art motion generation methods.

\end{abstract}


\IEEEpeerreviewmaketitle

\section{Introduction}
\label{sec:intro}

Generating realistic human motions is essential for human behavior understanding, since it can be applied for numerous applications including animation \cite{starke2021neural}, virtual or augmented reality \cite{petrovich2022temos,zhang2022wanderings}, as well as data-driven deep learning \cite{meng2019sample,zhao2020bayesian,liu2017enhanced,liu2019joint}. Despite the substantial progress made in modeling human dynamics, synthesizing natural, diverse, and controllable pose sequences remains challenging due to the complicated inter- and intra-class variations \cite{zhao2020bayesian}. 
Motion synthesis methods can be roughly divided into two categories: 1) unconditioned generation \cite{yan2019convolutional,zhao2020bayesian}  and 2) conditioned synthesis with the condition coming from various modalities \emph{i.e.}, music \cite{li2021ai}, speech \cite{ginosar2019learning,bhattacharya2021speech2affectivegestures}, emotion \cite{hou2020soul}, pre-defined action categories \cite{petrovich2021action,cervantes2022implicit}, text prompts \cite{guo2022action2video,petrovich2022temos}, past and/or future pose \cite{yuan2020dlow,harvey2020robust,duan2021single}.
In this paper, we focus on the conditioned generation case, aiming to generate more diverse 3D skeleton-based motion sequences in a controllable manner.

During the past decades, generative models for human motion generation have been investigating widely~\cite{barsoum2018hp,petrovich2021action,gui2018adversarial}. As a result, numerical methods were proposed based on Generative Adversarial Networks (GANs) \cite{barsoum2018hp,gui2018adversarial,harvey2020robust,davydov2022adversarial}, Variational Auto Encoder (VAE) \cite{pavlakos2019expressive,yuan2020dlow,petrovich2021action,petrovich2022temos}, Neural Distance Fields (NDF) \cite{peng2021neural,tiwari2022pose,cervantes2022implicit} or diffusion models \cite{kim2022flame,raab2022modi,findlay2022denoising,zhang2022motiondiffuse,tevet2022human}.
Generally, realism and diversity are two crucial aspects for evaluating the quality of the motion generation task.
To maintain diversity and boost realism during high-quality sampling, many approaches introduce GANs~\cite{goodfellow2020generative} to motion modality since GANs already showed remarkable success in image synthesizing. For instance, prior works \cite{barsoum2018hp} and \cite{gui2018adversarial} treated the predictor as a conditional generator and train the model in an adversarial manner with customized discriminators designed for human motion generation. Moreover, Harvey \emph{et al.}\cite{harvey2020robust} modified motion predictors into transition generators to tackle the in-between task, especially for variable-length complement given sparse key frames of animation. 

To produce a diverse set of possible future poses, Yuan \emph{et al.}~\cite{yuan2020dlow} proposed a new sampling strategy by exploiting conditional variational autoencoder (CVAE)~\cite{kingma2013auto} and KL constraints to balance between diversity and likelihood. Petrovich \emph{et al.}~\cite{petrovich2021action} first designed a Transformer-based CVAE for action-conditioned motion generation, namely ACTOR, which generates motions by sampling from the latent space based on parametric SMPL~\cite{loper2015smpl} model. More recently, they further extended ACTOR to text-conditioned motion generation~\cite{petrovich2022temos} by learning a joint latent space of two modalities \emph{i.e.}, motion and text based on a VAE framework.

The recent success of Neural Fields has attracted large interest in the computer vision community due to their effective capacity of representing complex data. Peng \emph{et al.}~\cite{peng2021neural} presented an implicit neural representation for dynamic humans for the new-view RGB video synthesis and 3D human model reconstruction. 
Unlike VAE-based approaches that infer variational distributions with an encoder, Cervantes \emph{et al.}\cite{cervantes2022implicit} proposed a model for action-conditional human motion generation that exploits variational Implicit Neural Representations (INR). 
In particular, their INR-based framework ensures sample-level representation optimization and enables variable-length sequence generation. Tiwari \emph{et al.}~\cite{tiwari2022pose} proposed to learn a human action prior that models the manifold of possible human poses represented by a scalar neural distance field, and employed this prior to downstream tasks \emph{i.e.}, denoising mocap data, occlusion data recovery, and 3D reconstruction from images. As a result, their formalism can project the human poses to a pose manifold instead of the Gaussian distribution that is commonly adopted in the VAE-based methods. This leads to a distance-preserving representation between poses.

\begin{figure*}[t] \small
	\centering
	\includegraphics[width=1\linewidth]{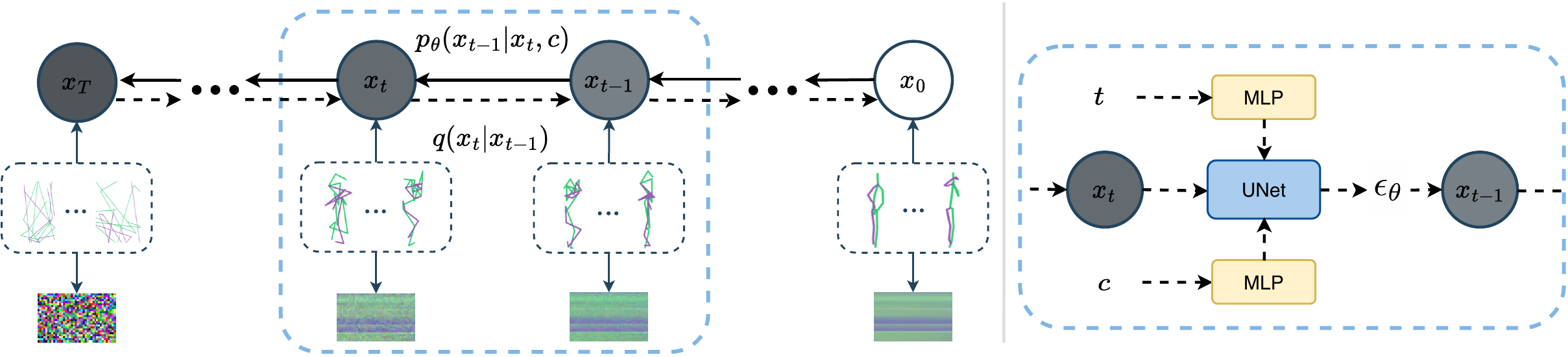}
	\caption{\textbf{Overview of Modiff}. The \textbf{diffusion} and \textbf{reverse} processes are represented by the solid and dashed arrows in the left part, respectively. During the reverse process, the detailed denoising procedure is exhibited in the right part. Note that, as shown in the bottom part, we treat the 3D skeleton sequence as a 2D image during training, and we also convert the sampled skeleton sequences into images to have a quick glance at the sampling results.}
	\label{fig:1} \vspace{-0.5em}
\end{figure*}

More recently, the diffusion model has shown its effectiveness in image synthesis~\cite{ho2020denoising,nichol2021improved,dhariwal2021diffusion}. Followed by these advances, several works have explored the capacity of the diffusion model for the human motion field \cite{kim2022flame,zhang2022motiondiffuse,tevet2022human}. These works investigated text-to-motion generation with a Transformer-based diffusion model on a large-scale dataset HumanML3D \cite{guo2022generating}. However, based on the observation of the generated results from the corresponding official implementation of \cite{tevet2022human},
we identify two limitations, \emph{i.e.}, 1) When an input text involves multiple actions, the generated sequences may fail to combine the corresponding actions properly or produce tangled motions with other unrelated actions. 2) Given a textual description with high-level semantic information \emph{e.g.} ``A person jumps up and down for two times”, their model generates incorrect 
actions, which means their model is not aware of the actual meaning of ``two times". 
In addition, Findlay \emph{et al.}~\cite{findlay2022denoising} employed DDPM \cite{ho2020denoising} for styled walking generation. Meanwhile, Saadatnejad \emph{et al.}~\cite{saadatnejad2022generic} introduced a framework for 3D human pose forecasting. It can produce future sequences recursively and incorporate denoising and post-processing by fully-exploit the properties of the diffusion model. However, their method is not an end-to-end paradigm, which increases the complexity of the task. For conditional action generation, \cite{zhang2022motiondiffuse,tevet2022human} applied their framework on the HumanAct12 and UESTC datasets, which contain 12 and 40 action classes, respectively. 
But for more challenging scenarios \emph{e.g.}, having more action categories to learn, the state-of-the-art approach is a GAN-based method \emph{i.e.} Kinetic-GAN~\cite{degardin2022generative}, which leverages GANs and Graph Convolutional Networks (GCNs) to generate human action sequences conditioned on the class label from Gaussian noise.

In this work, inspired by the diffusion model in image synthesis which outperforms GANs \cite{dhariwal2021diffusion}, we aim to explore the effectiveness of diffusion models on sequential skeleton-based motion generation. 
In particular, we focus on the problem of how to represent a 3D skeleton sequence and exploit the DDPM controlled by an action label. Conclusively, our goal is to take a pre-defined action label \emph{e.g.}, ``jump up” as input and generate an arbitrary number of plausible 3D human motion sequences via diffusion model.
In concentrate, our contribution is a conditional diffusion model-based motion generation framework. Specifically, we first treat the skeleton sequence as a 2D image matrix with 3 channels, then inject the label guidance into each reverse step during training and sampling. Finally, we evaluate our method on the challenging NTU RGB+D dataset with 60 action classes and the NTU RGB+D two-person dataset with 11 interaction classes. The experimental results show the proposed Modiff achieves state-of-the-art performances on both datasets.

\section{Diffusion Model for 3D Skeleton-Based Motion Generation}
\label{sec:method}

In this work, our goal is to tackle the task of action-conditioned 3D  skeleton-based human motion generation. Formally, given an action label $c \in\mathbb{C}$, where $\mathbb{C}$ is a predefined category set, we aim to generate arbitrary \emph{i.e.}, $N$ sequences of body poses $x^N_{1:f}$ with $f$ frames. In our case, each pose $x_{i}\in\mathbb{R}^{3 \times J}$ corresponds to an articulated human body, \emph{i.e.}, the 3D coordinates of human joints, where $J$ represents the number of joints. We use $x$ as a shorthand for each motion sequence $x_{1:f}$. Hence, our motion diffusion process operates on a sequence level.

\subsection{Denoising diffusion probabilistic model}
As shown in Figure \ref{fig:1}, diffusion model\cite{sohl2015deep} is composed of two mapping processes using the Markov chain. Each process essentially converts one distribution \emph{e.g.}, Gaussian into another \emph{e.g.}, motion data. According to the starting and endpoint of Markov chains, the two processes are named forward diffusion process for the inference and reverse diffusion process for the generation. 

Given the real (motion) data drawn from the distribution $x_0 \sim q\left(x_0\right)$ with dimension ${D}$, the \textit{diffusion process} is producing the latent $x_{t}\in\mathbb{R}^{D}$ by adding Gaussian noise to the data progressively at time $t\in\left\{1,\cdots, T\right\}$ :
\begin{equation}
\begin{aligned}
q\left(x_{0:T}\right)
&:=q\left(x_{0}\right) \prod^{T}_{t=1} q\left(x_t \mid x_{t-1}\right), \\
q\left(x_{1:T} \mid x_0\right) &:=\prod^{T}_{t=1} q\left(x_t \mid x_{t-1}\right), \\
q\left(x_t \mid x_{t-1}\right) &:=\mathcal{N}\left(x_t ; \sqrt{1-\beta_t} x_{t-1}, \beta_t \textbf{I}\right),
\end{aligned}
\end{equation}
where $\beta_t \in(0,1)$ is the variance of Gaussian noise. Symmetrically, the \textit{reverse process} is as follows:
\begin{equation}
\begin{aligned}
p_\theta\left({x}_{0: T}\right)
&:=p\left({x}_T\right) \prod_{t=1}^T p_\theta\left({x}_{t-1} \mid {x}_t\right), \\
p_\theta\left(x_{t-1} \mid x_t\right)
&:=\mathcal{N}\left(x_{t-1} ; \mu_\theta\left(x_t, t\right), \Sigma_\theta\left(x_t, t\right)\right),
\end{aligned}
\end{equation}
where $p\left(x_T\right)=\mathcal{N}\left(0, \mathbf{I}\right)$. Notably, this posterior can be used to generate sample $x_0$ given a Gaussian noise $x_T \sim \mathcal{N}\left(0, \mathbf{I}\right)$ and reducing the noise gradually from time step $T$ to $0$. Moreover, $\mu_\theta$ and $\Sigma_\theta$ are derived from the diffusion step $t$ and latent $x_t$, which describe the learned Gaussian transition by the model after training to remove the Gaussian noise from $x_t$ to $x_{t-1}$.
Thus, sample $x_{t-1} \sim p_\theta\left({x}_{t-1} \mid {x}_t\right)$ is to calculate:

\begin{equation}
x_{t-1} = \frac{1}{\sqrt{\alpha_t}}\left(x_t-\frac{\beta_t}{\sqrt{1-\bar{\alpha_t}}}\epsilon_\theta\left(x_t,t\right)\right)+\sigma_t\mathbf{z},
\end{equation}
where $\mathbf{z} \sim \mathcal{N}=\left(0,\mathbf{I}\right)$ and $\sigma_t \approx \sqrt{\beta_t}$, which means $\Sigma_\theta\left(x_t, t\right)=\sigma^2_t\mathbf{I}$ is fixed to a constant.  

\begin{algorithm}[t]
	\renewcommand{\algorithmicrequire}{\textbf{Input:}}
	\renewcommand{\algorithmicensure}{\textbf{Output:}}
	\caption{Action-conditioned Sampling Algorithm}
	\begin{algorithmic}[1]
		\REQUIRE Action label $c$, learned $\epsilon_\theta\left(x_t,t,c\right)$
		\ENSURE Motion sequence $x_0$
		\STATE 
		$x_T \leftarrow$ sample from $p\left(x_T\right)=\mathcal{N}\left(0, \mathbf{I}\right)$
		\STATE 
		\textbf{for} t = T, T-1, \dots, 1 \textbf{do}
		\STATE 
		~~~~$\mu_\theta\left(x_t,t,c\right) \leftarrow$ calculate from the learned $\epsilon_\theta\left(x_t,t,c\right)$ using Eq.\ref{eq:7}
	    \STATE
	    ~~~~$x_{t-1} \leftarrow$ sample from $\mathcal{N}\left(x_{t-1} ; \mu_\theta\left(x_t, t, c\right), \sigma^2_t\textbf{I}\right)$
		\STATE
		\textbf{end for}
		\STATE 
		\textbf{return} $x_0$
	\end{algorithmic}  
	\vspace{-0.2em}
\label{alg:2}
\end{algorithm}

During the training phase, \cite{ho2020denoising} proposed to sample latent $x_t$ from arbitrary step $t$ with $x_0$. To be specific, we sample step $t \sim [1, T]$ from a uniform distribution and get intermediate $x_t$ from $q\left(x_t \mid x_0\right)$ as follows: 
\begin{equation}
\begin{aligned}
    q\left(x_{t} \mid x_0\right) &=\mathcal{N}\left(x_t; \sqrt{\bar{\alpha_t}}x_0,\left(1-\bar{\alpha_t}\right)\textbf{I}\right), \\
    x_t
    &=\sqrt{\bar{\alpha}_t} x_0+\sqrt{1-\bar{\alpha}_t} \epsilon
\end{aligned}
\end{equation}
where $\bar{\alpha_t}=\prod^{t}_{m=0}\alpha_m$ and $\alpha_t=1-\beta_t$, $\epsilon \sim \mathcal{N}({0},\textbf{I})$. 
While \cite{nichol2021improved} found that the linear noise schedule is not suitable for low-resolution images, since skeleton data is rather spatially-sparse which can be treated as low-resolution images, we adopt the noise schedule in \cite{nichol2021improved}:
\begin{equation}
\bar{\alpha}_t=\frac{f(t)}{f(0)}, \quad f(t)=\cos \left(\frac{t / T+m}{1+m} \cdot \frac{\pi}{2}\right)^2,
\end{equation}
where $s=0.008$. Hence, $\beta_t=1-\frac{\bar{\alpha}_t}{\bar{\alpha}_{t-1}}$. We feed $t$ and $x_t$ to a U-Net\cite{ronneberger2015u} model as \cite{ho2020denoising}, and predict the $\epsilon_\theta$ by minimizing the L1 loss experimentally:
\begin{equation}
L_{\text {noise}}=E_{t, x_0, \epsilon}\left[\lvert \epsilon-\epsilon_\theta\left(x_t, t\right)\rvert\right].
\end{equation} 
Therefore, $\mu_\theta\left(x_t,t\right)$ can be derived from $\epsilon_\theta\left(x_t, t\right)$ as:
\begin{equation}
\mu_\theta\left(x_t, t\right)=\frac{1}{\sqrt{\alpha_t}}\left(x_t - \frac{\beta_t}{\sqrt{1-\bar\alpha_t}}\epsilon_\theta\left(x_t, t\right) \right).
\label{eq:7}
\end{equation}

\subsection{Action-conditioned diffusion}
\label{sec:label}
We leverage a classifier-free diffusion \cite{ho2022classifier} for action-conditioned motion generation, which means no extra classifier is required at the training stage. The label of action class $c$ is added in the denoising process, which is formulated as:
\begin{equation}
p_\theta\left(x_{t-1} \mid x_t, c\right):=\mathcal{N}\left(x_{t-1} ; \mu_\theta\left(x_t, t, c\right), \Sigma_\theta\left(x_t, t, c\right)\right).
\end{equation}
In the implementation, we inject class prior by adding a class embedding in the same way as the time step \cite{nichol2021improved} as described in Figure \ref{fig:1}. 
The summary of the sampling algorithms for action-conditioned diffusion is given in Algorithm \ref{alg:2}.

\section{Experiments}
\label{sec:experi}

\subsection{Datasets \& Settings}
\noindent \textbf{NTU RGB+D dataset~\cite{shahroudy2016ntu}.} This is a widely used dataset for skeleton-based action recognition with 3D skeleton annotations. It is composed of $56,880$ training samples for $60$ action classes which are collected from $40$ human subjects.

\noindent \textbf{NTU RGB+D two-person dataset~\cite{shahroudy2016ntu}.} We use 11 categories which contain the interactions of two people \emph{e.g.}, ``Kicking", ``Hugging" and ``Pushing" for state-of-the-art comparison.

\noindent \textbf{Evaluation metrics.} We evaluate our method on the cross-subject benchmark. This dataset is divided into two subsets according to the IDs of subjects. Consequently, the training set contains $40,320$ action sequences, and the testing set contains $16,560$ action sequences.
To evaluate the quantity of generated motion sequence, we adopt Fr\'echet Motion Distance (FMD) proposed in \cite{park2021diverse} as a quantitative metric for computing the distance between the two feature distributions of the real and generated motion samples.
We also follow \cite{petrovich2021action}, measure the overall diversity and intra-class diversity (denoted as multimodality).
Our method is implemented with PyTorch framework and is trained for 100 epochs using the Adam optimizer \cite{kingma2014adam} on one Nvidia RTX A6000 GPU. Before training, we uniform the 3D coordinates of human joints during preprocessing.

\renewcommand\arraystretch{1.5}
\begin{table}[t]
\centering
\caption{The results of our Modiff on the NTU RGB+D dataset. We compare our model with Kinetic-GAN~\cite{degardin2022generative} on cross-subject benchmark. (Best results in bold)}
\label{tab1}
\resizebox{31em}{!}{
\begin{tabular}{l|c|c|c}
\hline
\multicolumn{1}{c|}{Method}& FMD$\downarrow$ & Diversity$\uparrow$ & Multimodality$\uparrow$\\ \hline
Real          & 0.39$^{\pm0.00}$     & 36.90$^{\pm0.13}$  & 26.79$^{\pm0.04}$ 
\\ \hline
Kinetic-GAN-MLP8~\cite{degardin2022generative}   & 82.88$^{\pm8.14}$    & 4.47$^{\pm0.12}$    &  3.97$^{\pm0.07}$ \\
Kinetic-GAN-MLP4~\cite{degardin2022generative}   & 9.99$^{\pm2.31}$      & $12.83^{\pm0.20}$   & $11.04^{\pm0.09}$ \\
\rowcolor{blue!5} Modiff (Ours) & \textbf{9.12}$^{\pm0.00}$  & 
\textbf{12.85}$^{\pm0.19}$  & \textbf{11.48}$^{\pm0.08}$  \\ \hline
\end{tabular}}
\label{tab:fmd}\vspace{-1em}
\end{table}

\begin{figure}[t] \small
	\includegraphics[width=1\linewidth]{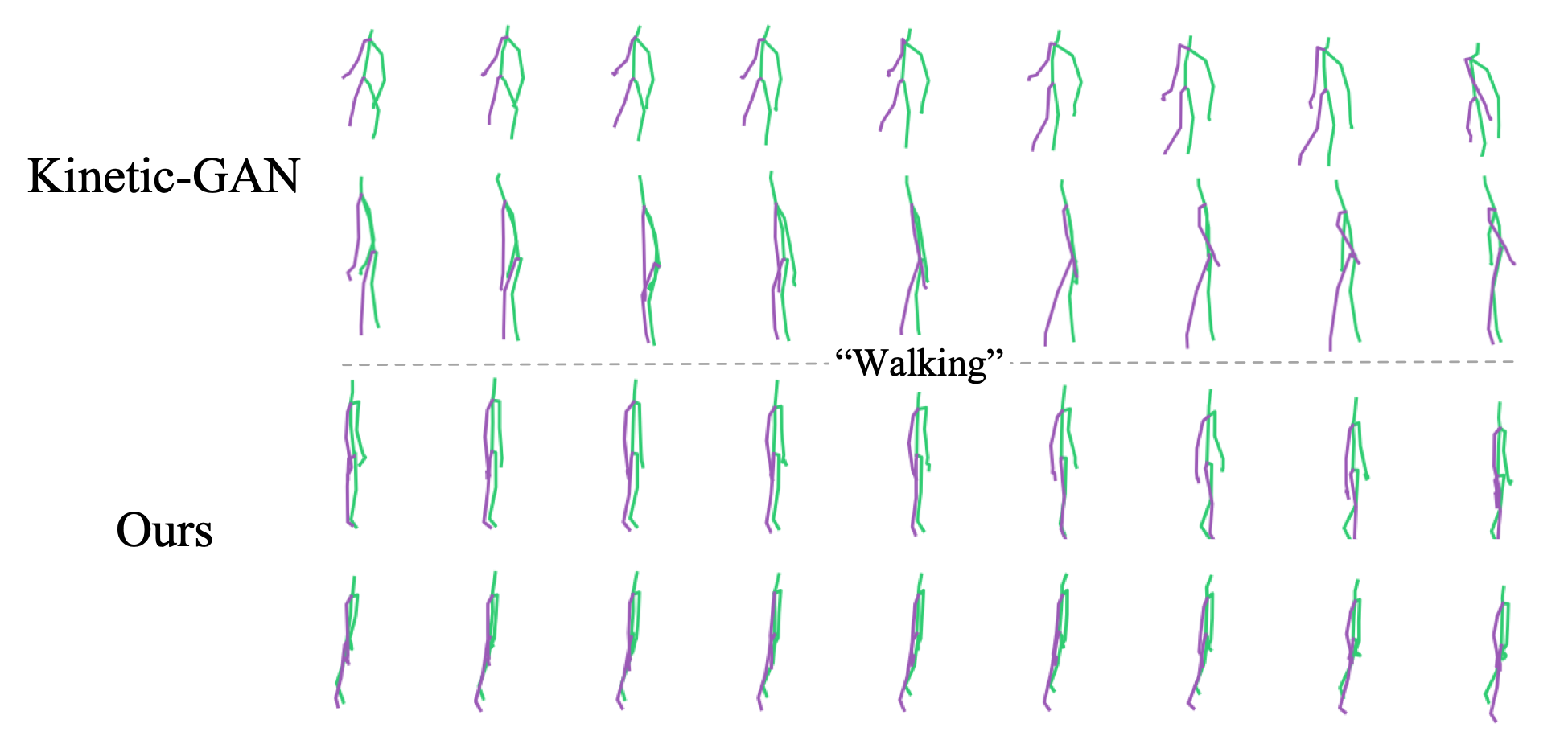}
	\caption{Visualization of generated motion sequence from Kinetic-GAN~\cite{degardin2022generative} and Ours by given label ``Walking” on the NTU RGB+D dataset.}
\label{fig:ours-vs-KG}
\end{figure}

\begin{table}[!ht]
\centering
\caption{The results of our Modiff on the NTU RGB+D two-person dataset. We compare our model with Kinetic-GAN$^\star$~\cite{degardin2022generative} on cross-subject benchmark. (Best results in bold)}
\resizebox{31em}{!}{
\begin{tabular}{l|c|c|c}
\hline
\multicolumn{1}{c|}{Method}& FMD$\downarrow$ & Diversity$\uparrow$ & Multimodality$\uparrow$\\ \hline
Real          & 1.79$^{\pm0.00}$     &  22.65$^{\pm0.21}$  &  20.72$^{\pm0.23}$ 
\\ \hline
Kinetic-GAN$^\star$~\cite{degardin2022generative}   & 140.49$^{\pm0.00}$      & $5.04^{\pm0.03}$   & $4.62^{\pm0.08}$ \\
\rowcolor{blue!5} Modiff (Ours) & \textbf{76.00}$^{\pm0.57}$  & 
\textbf{21.75}$^{\pm0.15}$  & \textbf{19.69}$^{\pm0.52}$  \\ \hline
\end{tabular}}
\label{tab:fmd2person}\vspace{-1em}
\end{table}

\begin{figure}[t] \small
    \centering
	\includegraphics[width=1.0\linewidth, height=8.5cm]{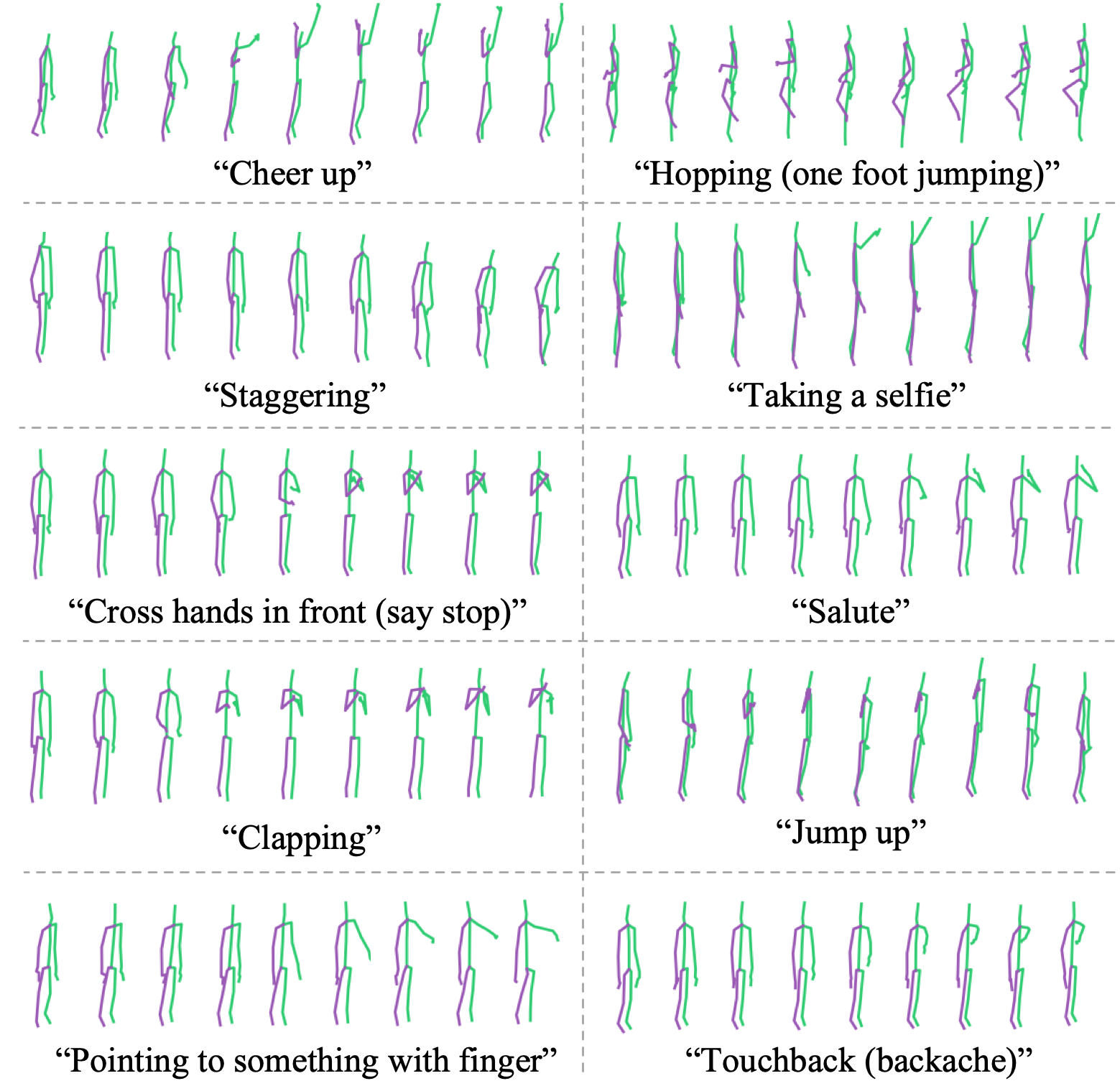}
	\caption{Visualization of generated motion sequences of our method on the NTU RGB+D dataset. (Best viewed when zoomed in)}
\label{fig:ours}\vspace{-1em}
\end{figure}

\subsection{Comparison with State-of-the-art Methods}
In Table \ref{tab:fmd}, we compare our method with the previous state-of-the-art approach Kinetic-GAN~\cite{degardin2022generative}. We report the data by using their pre-trained models with released code for a fair comparison. Our results outperform Kinetic-GAN across all types of metrics including FMD, Diversity, and Multimodality. Note that, our presented method is far better than Kinetic-GAN-MLP8 in terms of mean FMD (9.89 vs 82.88). In Table \ref{tab:fmd2person}, we adapt the code of Kinetic-GAN~\cite{degardin2022generative} for two-person action generation with slight modification, denoted as Kinetic-GAN$^\star$. Our presented method clearly surpasses Kinetic-GAN$^\star$ on all the metrics. In particular, we achieve samples with lower FMD, which is 64.49 less than Kinetic-GAN$^\star$.

We further illustrate the comparison of qualitative results in Figure \ref{fig:ours-vs-KG} between kinetic-GAN and ours on ``Walking". We sampled the same number of sequences from ours and kinetic-GAN and choose the best two for comparison. The first motion sequence produced by kinetic-GAN has almost no walking movement, the synthesized person just stands still with the left arm and leg moving slightly over the time span. Also, the second sample shows an unrealistic human motion dynamic, which swaps the left and right part of the body and almost without alternate footsteps. By contrast, our results are more robust and diverse in the rhythm of paces. Although our Modiff shows the promising performance of generating sequential motion samples conditioned on the action type, it still has limitations. 1) Similar to Kinetic-GAN~\cite{degardin2022generative}, ACTOR~\cite{petrovich2021action}, and other action-conditioned generation methods, the joints in the produced motion sequence may exist jitters over time. 2) Similar to Kinetic-GAN~\cite{degardin2022generative}, a small portion of sampling results may crash from the beginning or crash at the end.

\begin{table}[t]
    \centering
    \caption{Ablation study results of U-Net, loss, and label embedding.}
    \resizebox{0.485\textwidth}{!}{
    \begin{tabular}{ccc|cc|ccc|c}
    \cmidrule(r){2-9}
    & Label emb.\cite{tevet2022human}  & Label emb.\cite{nichol2021improved} & L2 loss  & L1 loss  & U-Net-16 & U-Net-32 & U-Net-64 & FMD$\downarrow$ \\ \cmidrule(r){2-9}
    B1 & \checkmark   &      & \checkmark &   & \checkmark   &  &  {}   & 161.44$^{\pm0.31}$\\ \cmidrule(r){2-9}
    B2 & \checkmark   &      & \checkmark &   &    & \checkmark  &  {}   & 29.11$^{\pm0.42}$\\ \cmidrule(r){2-9}
    B3 & \checkmark   &      & \checkmark &   &    &   &  \checkmark    & 15.70$^{\pm1.23}$ \\ \cmidrule(r){2-9}
    B4 & \checkmark   &      &  & \checkmark  &    &    &  \checkmark    & 9.42$^{\pm0.00}$ \\ \cmidrule(r){2-9}
    B5 &    &    \checkmark  &  & \checkmark  &    &    &  \checkmark    & \textbf{9.12}$^{\pm0.00}$  \\ \cmidrule(r){2-9}
    B6 &    &    \checkmark  & \checkmark  &  &    &    &  \checkmark    &  14.34$^{\pm0.31}$  \\ 
    \cline{2-9}
    \end{tabular}}
    \label{tab:abla}\vspace{-1em}
\end{table}

\subsection{Ablation Study}
We provide the qualitative results in Figure \ref{fig:ours} on a bunch of action types, including ``Cheer up”, ``Hopping”, ``Taking a selfie”, ``Staggering”, etc. As seen in the generated samples, our method can capture the underlying pattern of predefined actions and produce a realistic sequence with representative keyframes. For instance, one can easily recognize the ``Salute” action according to the synthesized movement of raising one hand and placing it next to the ear. Similarly, it's also intuitive for the ``Cross hands in front” sample when we see the generated person cross hands over the chest.

\noindent \textbf{Effect of U-Net.} Moreover, as illustrated in Table \ref{tab:abla}, we offer the quantitative results of U-Net with different dimensions of latent layers. U-Net-16 means for up and down layers, the minimum dimension is 16, the same as the U-Net-32 and U-Net-64. Intuitively, comparing baseline B3 with B1 and B2, U-Net with higher dimension layers leads to effective latent representation for motion data and achieves better generation results. B3 outperforms B1 and B2 by a large margin on FMD, which decrease FMD metric by 145.74. 

\noindent \textbf{Effect of Loss.} As observed in Table \ref{tab:abla}, we also report the quantitative results of different losses. According to the results of metrics from baseline B3, B4, B5, and B6, L1 loss results in an improvement of 6.28 (B3 vs B4) and 5.22 (B5 vs B6) on FMD metric.

\noindent \textbf{Effect of Label embedding.} As indicated in Table \ref{tab:abla}, we investigate two different ways of label embedding. Label emb.\cite{nichol2021improved} means the method introduced in Section \ref{sec:label}, Label emb. \cite{tevet2022human} means the approach used in \cite{tevet2022human}. By comparing baseline B4 with B5 (our Modiff), Label emb.\cite{nichol2021improved} further boosts the performance of our Modiff by 0.3 on FMD.

\section{Conclusion}
\label{sec:conclusion}

This work introduces Modiff, a novel framework based on the diffusion probabilistic model for action-conditioned motion generation. More specifically, the presented Modiff integrates the denoising diffusion model with motion representations, injecting the action-type condition to the reverse Markov chain for action-conditioned motion synthesis. We conducted extensive experiments on a large-scale NTU RGB+D dataset and achieved competitive qualitative and quantitative results compared to the state-of-the-art motion generation method.



\clearpage
\bibliographystyle{IEEEtran}
\bibliography{IEEEabrv,ref}
 
\end{document}